\documentclass[preprint,12pt,3p,review]{elsarticle}

\usepackage{amssymb,amsmath}
\usepackage{color}
\usepackage{multirow}
\usepackage{booktabs}
\usepackage{siunitx}
\usepackage{soul}
\usepackage{array}
\newcolumntype{P}[1]{>{\centering\arraybackslash}p{#1}}
\newcolumntype{M}[1]{>{\centering\arraybackslash}m{#1}}
\usepackage{longtable}
\usepackage{lscape}
\usepackage{setspace}
\usepackage{graphicx}
\usepackage{setspace}
\usepackage[numbers]{natbib}
\usepackage{subcaption}
\usepackage[utf8]{inputenc}
\usepackage[ruled]{algorithm2e}
\usepackage{caption}
\usepackage{makecell}
\usepackage{float}
\usepackage[usenames,dvipsnames]{xcolor}
\usepackage{tcolorbox}
\usepackage{tabularx}
\usepackage{array}
\usepackage{colortbl}
\usepackage{geometry}
\usepackage[hyphens]{url}
\usepackage{xcolor}
\usepackage{url}
\usepackage{wrapfig}
\usepackage[colorlinks = true,
            linkcolor = blue,
            urlcolor  = blue,
            citecolor = blue,
            anchorcolor = blue,
            pdfauthor=author]{hyperref}
\emergencystretch=\maxdimen
\tcbuselibrary{skins}

\newcolumntype{Y}{>{\raggedleft\arraybackslash}X}

\usepackage{bm,graphicx}
\usepackage{amsmath}
\usepackage{fullwidth}
\usepackage{subcaption}
\newenvironment{walgorithm}{
\centering
\begin{minipage}{0.95\linewidth}
\begin{algorithm}[H]}{\end{algorithm}
\end{minipage}
}

\DeclareMathOperator*{\argmax}{arg\,max}
\def\tsc#1{\csdef{#1}{\textsc{\lowercase{#1}}\xspace}}
\tsc{WGM}
\tsc{QE}
\tsc{EP}
\tsc{PMS}
\tsc{BEC}
\tsc{DE}
\graphicspath{{Images/}}

\begin{document}

\begin{frontmatter}
\title{IKDSumm: Incorporating Key-phrases into BERT for extractive Disaster Tweet Summarization}
\author[a]{Piyush Kumar Garg\corref{cor1}}
\ead{piyush_2021cs05@iitp.ac.in}

\author[b]{Roshni Chakraborty}
\ead{roshni.chakraborty@ut.ee}

\author[a]{Srishti Gupta}\ead{srishti_2021cs38@iitp.ac.in}

\author[a]{Sourav Kumar Dandapat}
\ead{sourav@iitp.ac.in}


\address[a]{Department of Computer Science \& Engineering, Indian Institute of Technology Patna, Bihar, India}

\address[b]{Institute of Computer Science, University of Tartu, Estonia}

\begin{abstract}
Online social media platforms, such as Twitter, are one of the most valuable sources of information during disaster events. Therefore, humanitarian organizations, government agencies, and volunteers rely on a summary of this information, i.e., tweets, for effective disaster management. Although there are several existing supervised and unsupervised approaches for automated tweet summary approaches, these approaches either require extensive labeled information or do not incorporate specific domain knowledge of disasters. Additionally, the most recent approaches to disaster summarization have proposed BERT-based models to enhance the summary quality. However, for further improved performance, we introduce the utilization of domain-specific knowledge without any human efforts to understand the importance (salience) of a tweet which further aids in summary creation and improves summary quality. In this paper, we propose a disaster-specific tweet summarization framework, \textit{IKDSumm}, which initially identifies the crucial and important information from each tweet related to a disaster through key-phrases of that tweet. We identify these key-phrases by utilizing the domain knowledge (using existing ontology) of disasters without any human intervention. Further, we utilize these key-phrases to automatically generate a summary of the tweets. Therefore, given tweets related to a disaster, \textit{IKDSumm} ensures fulfillment of the summarization key objectives, such as information coverage, relevance, and diversity in summary without any human intervention. We evaluate the performance of \textit{IKDSumm} with $8$ state-of-the-art techniques on $12$ disaster datasets. The evaluation results show that \textit{IKDSumm} outperforms existing techniques by approximately $2-79\%$ in terms of ROUGE-N F1-score.

\end{abstract}

\begin{keyword} Social media \sep Disaster events \sep Tweet summarization \sep Key-phrase extraction
\end{keyword}
\end{frontmatter}

\section{Introduction} \label{s:intro}
\par Social media platforms, such as Twitter, have become important sources of information during disaster events~\cite{purohit2020ranking, priya2019should, ali2020topic}. For example, government agencies, humanitarian organizations, and volunteers rely on user-shared updates to ensure  effective assessment and immediate relief operations~\cite{castillo2016big, andrews2016creating}. However, the huge volume of user updates often makes it difficult for organizations to manually identify the required relevant information. Therefore, several research works have been proposed for summarization to automatically process the huge number of tweets into a concise summary~\cite{dutta2019community, sharma2019going, rudra2018identifying, wang2019microblog}.

\par Existing disaster-specific extractive summarization approaches are either supervised approaches~\cite{nguyen2022towards, li2021twitter, dusart2021issumset, rudra2016summarizing} or unsupervised approaches~\cite{Garg2022Entropy, dutta2018ensemble, dutta2019community, rudra2018identifying}. Irrespective of the approaches, most of the existing methods assess the importance of each tweet and finally select the most important tweets to create the summary.
Existing unsupervised approaches assess the importance of tweets based on the content words or context of the tweet. For example, Rudra et al.~\cite{rudra2015extracting} utilizes the content words (i.e., nouns, verbs, and adjectives), and Rudra et al.~\cite{rudra2018identifying} utilizes noun-verb pairs to identify the importance of the tweets related to a disaster event. However, these approaches do not exploit disaster-specific domain knowledge to determine the importance of tweets and hence fail to capture the real importance of tweets from the perspective of a disaster event. 

\par Existing supervised approaches, such as Dusart et al.~\cite{dusart2023tssubert} predict the importance of a tweet on the basis of the importance of a tweet with respect to the model by a BERT model~\cite{devlin2018bert}. However, they utilize context embedding of term-frequencies of tweets to determine tweets' importance, which does not pay more attention to  key-phrases that have disaster-specific keywords than other keywords in tweets. To handle this issue, Nguyen et al.~\cite{nguyen2022towards, nguyen2022rationale} highlight the significance of understanding the most important disaster-related keywords or key-phrases~\footnote{The short snippets in tweets which provide the important information in tweets, called key-phrases.} with respect to a disaster to determine the importance of a tweet. However, these approaches rely on supervised BERT-based multi-task learning approaches, which require a huge number of labeled tweets (with annotated key-phrase) that are very difficult and costly to obtain for disaster events. Therefore, although unsupervised approaches are cheaper, they do not consider the domain knowledge to determine the importance of a tweet. Although few supervised approaches consider the relevance of domain knowledge, these approaches are dependent on a reasonably sized labeled training set which makes these approaches expensive.

\par To get the best of both of these types of approaches, we propose a summarization approach \textit{IKDSumm}, which utilizes existing ontology knowledge instead of labeled training data to identify key-phrase and then to create a summary. For key-phrase extraction, we propose an unsupervised \textbf{D}isaster-specific \textbf{R}apid \textbf{A}utomatic \textbf{K}eyword \textbf{E}xtraction (DRAKE), which is embedded with disaster specific knowledge and a modified version of RAKE~\cite{rose2010automatic}. This key-phrase extraction methodology significantly reduces the cost of creating a summary without compromising summary quality. The novelty of DRAKE is that it utilizes disaster ontology knowledge to extract the key-phrases from tweets and, therefore, can ensure the maximum information coverage regarding disasters in key-phrases without any human intervention. \textit{IKDSumm} finally utilizes these extracted key-phrases to automatically generate a tweet summary which comprises of tweets with maximum information and minimum redundancy. To show the effectiveness of our approach, we evaluate the performance of \textit{IKDSumm} on $12$ disaster events and, further, compare our results with $8$ existing state-of-the-art research works in terms of ROUGE-N~\cite{lin2004rouge} scores. Additionally, we observe that DRAKE is better than the existing key-phrase extraction methods, such as RAKE, and Nguyen et al.~\cite{nguyen2022towards} by $4.02-27.83\%$ and $2.33-26.53\%$ in terms of Intersection-Over-Union (IOU) (F1-score) and Jaccard Similarity score, on $7$ disaster events  respectively. We release~\footnote{\url{https://drive.google.com/drive/folders/1GtwStIcyGXbKFNA9_K2TRmP11D1LpPeg?usp=sharing}} annotated key-phrases for $7$ disaster datasets of different locations and types.

\par The organization of the rest of the paper is as follows. We discuss related works in Section~\ref{s:rworks} and the dataset details in Section~\ref{s:data}. In Section~\ref{s:propApp}, we discuss our proposed approach. We discuss the experiment details in Section~\ref{s:exp} and results in Section~\ref{s:result}. Finally, we conclude our paper in Section~\ref{s:con}.

\section{Related Works} \label{s:rworks}
\par Summarization represents the main aspects and the key information of an input text such that it provides a concise overview of the text~\cite{saranyamol2014survey, zhao2021qbsum, canhasi2016weighted}. There are several applications of text summarization, such as legal texts summarization~\cite{bhattacharya2021incorporating, bhattacharya2019comparative}, news summarization~\cite{ahuja2022aspectnews, hernandez2022language, curiel2021online}, product review summarization~\cite{komwad2022survey}, tweet summarization~\cite{chakraborty2019tweet,chakraborty2017network}, timeline summarization~\cite{yu2021multi, ansah2019graph}, etc. Social media platforms, like Twitter~\footnote{https://en.wikipedia.org/wiki/Twitter}, have become extensively popular to share news, opinions and updates for ongoing events. However, a  huge surge of tweets on Twitter makes it challenging to identify the relevant information from tweets and as a result, tweet summarization got huge attention from the research community. There are several existing tweet summarization approaches, which could be categorized into sports event summarization~\cite{goyal2019multilevel, huang2018event, gillani2017post}, news event summarization~\cite{zheng2021tweet, duan2019across, chakraborty2017network}, political event summarization~\cite{panchendrarajan2021emotion, kim2014tweet}, disaster event summarization~\cite{saini2019multiobjective, saini2021microblog}, social event summarization~\cite{narmadha2016survey, schinas2015visual}, etc., on the basis of various domains. 

\par Several prior studies~\cite{roy2020classification, dutta2019summarizing, rudra2019summarizing} attempt to summarize disaster event tweets that comprise situational and time-critical information posted by eye-witnesses and affected people during a disaster. Existing disaster-specific tweet summarization approaches can be categorized into abstractive~\cite{vitiugin2022cross, nguyen2022towards, lin2021preserve, rudra2016summarize} or extractive summarization~\cite{Garg2022Entropy, saini2021microblog, dusart2021issumset, rudra2016summarize} approaches. In this paper, we focus on the extractive summarization approach, and therefore, we only discuss related works on the extractive disaster tweet summarization approaches next. Existing extractive disaster tweet summarization approaches are either graph-based approaches~\cite{dutta2018ensemble, dutta2015graph}, content-based approaches~\cite{rudra2018extracting, sharma2019going}, deep learning-based approaches approaches~\cite{li2021twitter, dusart2023tssubert} and an ontology-based approach~\cite{garg2023ontodsumm}.  

\par Existing content-based disaster tweet summarization approaches \cite{rudra2015extracting, rudra2018extracting} rely on identifying the importance of tweets on the basis of the frequency of the content words (i.e., nouns, verbs, and adjectives) to generate the summary. These approaches initially classify each tweet into either situational or non-situational tweets by semi-supervised learning~\cite{chen2015search} or supervised learning~\cite{roy2020classification, rudra2018classifying} and then select tweets that have higher importance into the summary. Additionally, some recent works~\cite{dusart2023tssubert, li2021twitter, zogan2021depressionnet} have proposed deep neural network architectures, such as disaster-specific BERT model~\cite{devlin2018bert}, the Graph Convolutional Neural network (GCN) based model~\cite{li2021twitter} to determine the tweet's importance based on various tweet content features and select the most important tweets to create a summary. However, these approaches require extensive labeled training data and human effort, which is very difficult to obtain in disaster scenarios.

\par In order to handle this, many researchers~\cite{dutta2018ensemble, dutta2019community, rudra2019summarizing} have proposed graph-based disaster tweet summarization approaches. These approaches initially construct a tweet-similarity graph where nodes represent the tweets and edges represent the similarity between tweets and then group similar tweets by identifying groups/communities. Finally, they select the representative tweets from each group/community on the basis of length, degree, and centrality-based methods to create a summary. For example, Dutta et al.~\cite{dutta2015graph, rudra2019summarizing} utilizes a community detection-based algorithm to identify groups of similar tweets and then select representative tweets from each community on the basis of the tweet length and degree-based measures. Thus, these approaches ensure coverage and diversity by selecting tweets from each community. However, automatic community identification is very challenging due to the high vocabulary overlap among tweets of different communities across disasters. Additionally, these approaches also do not consider the difference in importance of communities and their information content across disasters.

\par In order to resolve these challenges, Garg et al.~\cite{garg2023ontodsumm} proposed a tweet summarization framework, which initially identifies the category~\footnote{A category consists of information related to a same topic of a disaster.} of a tweet using an ontology-based pseudo-relevance feedback approach, followed by identification of the importance of each category which represents the number of tweets to be in summary from a category. Finally, they create a summary by selecting the representative tweets from each category by Disaster-specific Maximal Marginal Relevance (DMMR) based approach.  Recently, few approaches ~\cite{dusart2023tssubert, nguyen2022towards} have proposed utilization of BERT~\cite{devlin2018bert} based models for disaster summarization, such as Dusart et al.~\cite{dusart2023tssubert} predict the importance of a tweet by combining the context of the tweets with the vocabulary frequency related to the event and then, selecting the most important tweets to generate a summary. Similarly, Nguyen et al.~\cite{nguyen2022towards} utilize a supervised BERT-based approach to extract key-phrases from tweets and then create a summary by using an ILP-based optimization technique. However, to identify the key-phrases the supervised approaches require a huge number of labeled tweets for training which is very difficult to obtain in disaster scenarios. 
Therefore, in this paper, we propose \textit{IKDSumm} that initially identifies the key-phrases from each tweet using disaster domain knowledge from ontology, which does not need any human intervention, and then utilizes these key-phrases to predict the tweet salience score. Finally, we create the summary by iteratively selecting higher salience score tweets by maximizing the information content and diversity in the summary. We discuss datasets details next.   


\section{Dataset} \label{s:data}
\par In this Section, we discuss the datasets and the gold standard summary.  

\subsection{Dataset Details} \label{s:datadet}
\par We evaluate the performance of \textit{IKDSumm} on $12$ disaster datasets which are as follows. 

\begin{enumerate}
    \item \textit{$D_1$}: This dataset is created on the basis of the tweets related to the \textit{Sandy Hook Elementary School Shooting, United States}~\footnote{\url{https://en.wikipedia.org/wiki/Sandy\_Hook\_Elementary\_School\_shooting}} on December, $2012$ in which $26$ people were killed, including $20$ children and $6$ adults~\cite{dutta2018ensemble}.
    
    \item \textit{$D_2$}: This dataset is created on the basis of the tweets related to the \textit{Uttarakhand Flood, India}~\footnote{\url{https://en.wikipedia.org/wiki/2013\_North\_India\_floods}} on June, $2013$ in which $5,700$ people were killed, around $128$ houses were damaged, and more than $1,10,000$ people were evacuated~\cite{dutta2018ensemble}.
    
    \item \textit{$D_3$}: This dataset is created on the basis of the tweets related to the devastating impact of the terrible typhoon, \textit{Hagupit Typhoon, Philippines}~\footnote{\url{https://en.wikipedia.org/wiki/Typhoon\_Hagupit\_(2014)}} on December, $2014$ in which $18$ people were killed, around $916$ people were evacuated, and the estimated damage were around \$$114$ million USD~\cite{dutta2018ensemble}.  
    
    \item \textit{$D_4$}: This dataset is created on the basis of the tweets related to the \textit{Hyderabad Blast, India}~\footnote{\url{https://en.wikipedia.org/wiki/2013\_Hyderabad\_blasts}} on February, $2013$ in which $18$ people were killed including $2$ college students and around $131$ people were injured~\cite{dutta2018ensemble}. 
    
    \item \textit{$D_5$}: This dataset is created on the basis of the tweets related to the \textit{Harda Twin Train Derailment, India}~\footnote{\url{https://en.wikipedia.org/wiki/Harda\_twin\_train\_derailment}} on August, $2015$ in which around $31$ people were dead and around $100$ people were injured~\cite{rudra2018extracting}.  
    
    \item \textit{$D_6$}: This dataset is created on the basis of the tweets related to the \textit{Los Angeles International Airport Shooting, California}~\footnote{\url{https://en.wikipedia.org/wiki/2013\_Los\_Angeles\_International\_Airport\_shooting}} November, $2013$ in which $1$ person was killed and more than $15$ people were injured~\cite{olteanu2015expect}. 
    
    \item \textit{$D_7$}: This dataset is created on the basis of the tweets related to the \textit{Hurricane Matthew, Haiti}~\footnote{\url{https://en.wikipedia.org/wiki/Hurricane\_Matthew}} on October, $2016$ in which $603$ people were dead and around 128 people were missing~\cite{Alam2021humaid}.  

    \item \textit{$D_8$}: This dataset is created on the basis of the tweets related to the \textit{Puebla Mexico Earthquake, Mexico}~\footnote{\url{https://en.wikipedia.org/wiki/2017\_Puebla\_earthquake}} on September, $2017$ in which $370$ people were dead and more than $6000$ people were injured~\cite{Alam2021humaid}. 
    
    \item \textit{$D_9$}: This dataset is created on the basis of the tweets related to the devastating impact of the massive earthquake, \textit{Pakistan Earthquake, Pakistan}~\footnote{\url{https://en.wikipedia.org/wiki/2019\_Kashmir\_earthquake}} on September, $2019$ in which $40$ people were killed, around $850$ people were injured~\cite{Alam2021humaid}. 
    
    \item \textit{$D_{10}$}: This dataset is created on the basis of the tweets related to the \textit{Midwestern U.S. Floods, United States}~\footnote{\url{https://en.wikipedia.org/wiki/2019\_Midwestern\_U.S.\_floods}} on March, $2019$ in which around $14$ million people were affected and the estimated damage were around \$$2.9$ billion USD~\cite{Alam2021humaid}.

    \item \textit{$D_{11}$}: This dataset is created on the basis of the tweets related to the \textit{Kaikoura Earthquake, New Zealand}~\footnote{\url{https://en.wikipedia.org/wiki/2016\_Kaikoura\_earthquake}} on November, $2016$ in which $2$ people died, and $57$ got injured~\cite{Alam2021humaid}.

    \item \textit{$D_{12}$}: This dataset is created on the basis of the devastating impact of the strong cyclone, \textit{Cyclone Pam, Vanuatu}~\footnote{\url{https://en.wikipedia.org/wiki/Cyclone\_Pam}} in March, $2015$ which led to the death of around $15$ people and displacement of $3300$ people~\cite{imran2016twitter}.
\end{enumerate}

\begin{table*}[!ht]
    \caption{Table shows the details of $12$ disaster datasets, including dataset number, dataset name, number of tweets, summary length, country, and disaster type.}
    \label{table:dataset}
    \resizebox{\textwidth}{!}{\begin{tabular}{clcccc}
        \hline
        {\bf Num} & {\bf Dataset name} & {\bf Number of tweets} & {\bf Summary length} & {\bf Country} & {\bf Disaster type}\\ \hline
        
        $D_1$ & \textit{Sandy Hook Elementary School Shooting}      & 2080 & 36 tweets & USA            & Man-made   \\ 
        $D_2$ & \textit{Uttrakhand Flood}                           & 2069 & 34 tweets & India          & Natural  \\ 
        $D_3$ & \textit{Hagupit Typhoon}                            & 1461 & 41 tweets & Philippines    & Natural  \\ 
        $D_4$ & \textit{Hyderabad Blast}                            & 1413 & 33 tweets & India          & Man-made \\ 
        $D_5$ & \textit{Harda Twin Train Derailment}                & 4171 & 27 tweets & India          & Man-made \\ 
        $D_6$ & \textit{Los Angeles International Airport Shooting} & 1409 & 40 tweets & USA            & Man-made \\ 
        $D_7$ & \textit{Hurricane Matthew}                          & 1654 & 40 tweets & Haiti          & Natural  \\ 
        $D_8$ & \textit{Puebla Mexico Earthquake}                   & 2015 & 40 tweets & Mexico         & Natural  \\ 
        $D_9$ & \textit{Pakistan Earthquake}                        & 1958 & 40 tweets & Pakistan       & Natural  \\ 
     $D_{10}$ & \textit{Midwestern U.S. Floods}                     & 1880 & 40 tweets & USA            & Man-made \\ 
     $D_{11}$ & \textit{Kaikoura Earthquake} & 2195 & 40 tweets & New Zealand & Natural \\ 
     $D_{12}$ & \textit{Cyclone Pam} & 1508 & 40 tweets & Vanuatu & Natural \\ \hline
    \end{tabular}}
\end{table*}


\subsection{Gold Standard Summary} \label{s:groundtruth}
\par We use the gold standard summary provided by Dutta et al.~\cite{dutta2018ensemble} for $D_1$-$D_4$ datasets, Rudra et al.~\cite{rudra2018extracting} for $D_5$, and Garg et al.~\cite{garg2023ontodsumm} for $D_6$-$D_{12}$. We show the details of $D_1-D_{12}$ datasets and gold standard summary length in Table~\ref{table:dataset}.

\begin{figure}
    \centering 
    \includegraphics[width=15cm, height=10cm] {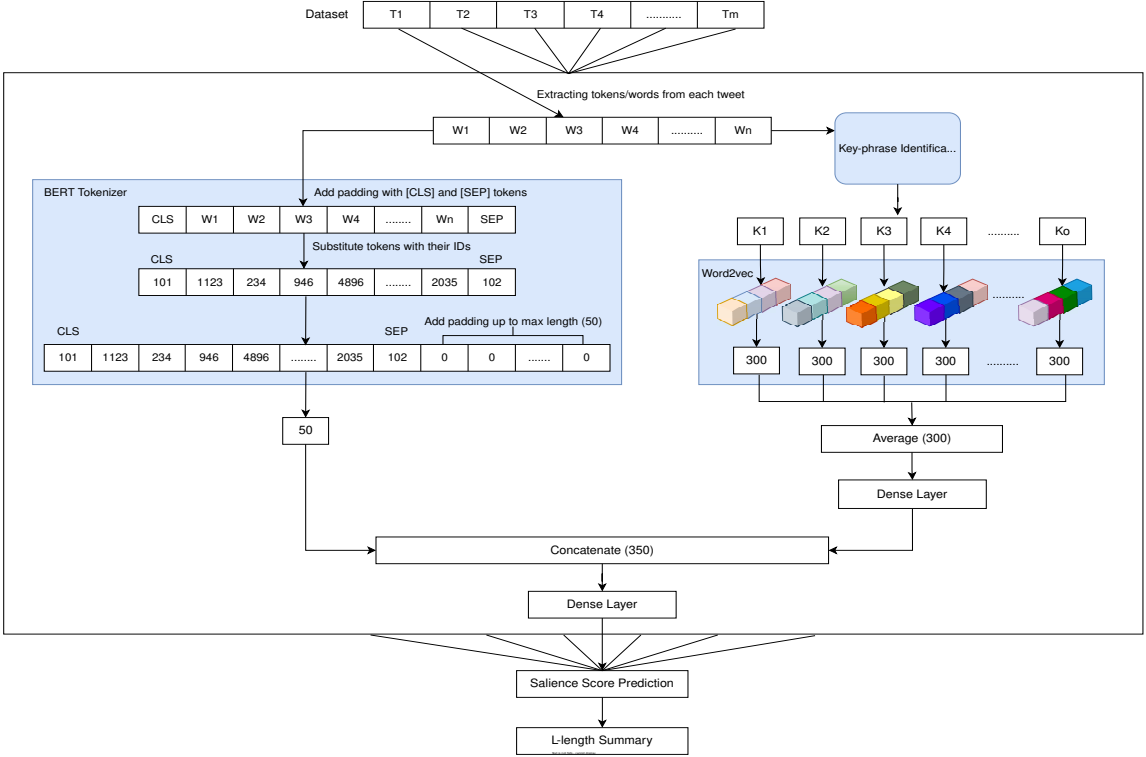}
    \caption{Architecture of \textit{IKDSumm}.}
    \label{figure:flowchart}
\end{figure}

\section{Proposed Approach} \label{s:propApp}
\par In this Section, we discuss \textit{IKDSumm} in detail. Given a disaster event, $D$, that consists of $m$ tweets, $T$ = \{$T_{1}, T_{2}, ..., T_{m}$\}, \textit{IKDSumm} generates a summary, $S$, by selecting $L$ tweets from $T$ such that it provides the maximum information coverage of $T$ and ensures maximum diversity in $S$. As in most summarization applications, we assume that the length of the summary, $L$, is provided. We show all the used notations and corresponding descriptions in Table~\ref{table:notation} and the architecture of \textit{IKDSumm} in Figure~\ref{figure:flowchart}, which comprises of the following components,  Tweet Encoder (Subsection \ref{s:tencod}), Key-phrase Extraction Approach (Subsection \ref{s:eea}), Key-phrase Encoder and Integrator (Subsection \ref{s:w2v}) and Summary Tweet Selector (Subsection \ref{s:sts}), discussed next in detail.

\subsection{Tweet Encoder} \label{s:tencod}
\par We generate the embedding of a tweet, $T_i$, which comprises of $n_i$ words, i.e., \{$W_1^{i}, W_2^{i}, ..., {W_{n_i}}^{i}$\} as $TE_{T_i}$ by a pre-trained DistilBERT model~\cite{sanh2019distilbert}, a lighter version of BERT~\cite{devlin2018bert}. For pre-processing of $T_i$, we append the special tokens, [CLS] and [SEP], to the start and at the end of $T_{i}$, respectively as \{[CLS],$W_1^{i}, W_2^{i}, ..., {W_{n_i}}^{i}$,[SEP]\}.  As the DistilBERT~\cite{sanh2019distilbert} model computation is quadratic to the length of the input, we fix the size to $50$ tokens to take advantage of the small size of a tweet (all of the tweets in our dataset have a length of $< 50$ tokens). If the tweet is smaller in size, we use the `PAD' token to reach the fixed limit size ($50$). We also add a token and replace tweet-specific instances, such as URLs by `url', hashtags by `htg', mentions (@) by `mtn', and retweets by 'rtw', respectively. 


\begin{table}
    \centering 
    \caption{Table shows the notations and their corresponding description used in \textit{IKDSumm}.}
    \label{table:notation}
    \begin{tabular}{cc}
        \hline
        {\bf Notation} & {\bf Description}  \\ \hline 
    
        $D$             & Dataset for a disaster event  \\
        $m$             & Number of tweets in $D$ \\
        $L$             & Desired length summary (number of tweets) \\
        $T$             & Set of tweets in $D$ \\  
        $S$             & Generated summary \\ 
        $T_i$           & $i^{th}$ indexed tweet in D \\ 
        $n_i$           & Number of words in $T_{i}$ of $D$\\ 
        $o_i$           & Number of \textit{Potential Key-phrases} in $T_{i}$ \\
        $TE_{T_i}$      & Tweet embedding of $T_{i}$ \\ 
        $KE_{T_i}$      & Key-phrase embedding of $T_{i}$ \\ 
        ${Kp}{_j}^{i}$  & $j^{th}$ key phrase of $T_{i}$ \\ 
        $StopW$         & List of stop words (or stoplist) \\ 
        $PhraseD$       & Set of phrase delimiters \\ 
        $WordD$         & Set of word delimiters \\ 
       $ContW(T_{i})$& \textit{Content Words} of $T_{i}$ \\  
$Sc(T_{i}, {Kp}{_j}^{i})$& Score of ${Kp}{_j}^{i}$ in $T_{i}$ \\ 
        $MaxSc({Kp}(T_i))$    & Highest \textit{Potential Key-phrase score} for $T_i$ \\ \hline
    \end{tabular}
\end{table}

\subsection{Key-phrase Extraction Approach} \label{s:eea}
\par While existing BERT-based supervised approaches~\cite{nguyen2022towards, nguyen2022rationale} require a huge number of labeled tweets to identify key-phrases, existing unsupervised methods, such as RAKE~\cite{rose2010automatic} do not require labeled tweets but fails to identify disaster specific keywords to identify key-phrases. Therefore, in \textit{IKDSumm}, we propose DRAKE, a modified version of RAKE specifically designed for disaster events, which provides higher importance to disaster-related words to extract key-phrases from tweets without any human intervention. 

\par Given a tweet $T_{i} \in T$, which comprises of $n_i$ words, stop words ($StopW$), phrase delimiters ($PhraseD$), and word delimiters ($WordD$), we identify the content words ($ContW(T_{i})$) after removing $StopW$ and $WordD$ from $T_{i}$ as shown in Equation~\ref{eq:contword}: 
\begin{align}
   ContW(T_{i}) = T_{i} - StopW - WordD
    \label{eq:contword}
\end{align}\

We identify the \textit{Potential Key-phrases} of $T_{i}$, i.e., $Kp(T_{i})$ = \{${Kp}{_1}^{i}, {Kp}{_2}^{i}, \ldots, {{Kp}_{o_i}^{i}}$\}, which are continuous sequences of content words (words belongs to $ContW (T_{i})$) in $T_{i}$. We calculate the score of $j^{th}$ \textit{Potential Key-phrase} of $T_i$, $Sc(T_{i}, {Kp}{_j}^{i})$ as the aggregation of its words, word-degree score, $S_{wd}(x)$ as shown in Equation~\ref{eq:contscor}: 

\begin{align}
   Sc(T_{i}, {Kp}{_j}^{i}) = \sum_{\forall x \in {Kp}{_j}^{i}}^{} S_{wd}(x)
    \label{eq:contscor}
\end{align}\

where $S_{wd}(x)$ is calculated based on the ratio of the frequency of a word ($freq(x)$) and the degree of correlation of the word ($deg(x)$) as shown in Equation~\ref{eq:wordscor}~\cite{rose2010automatic}. While $freq(x)$ is the frequency of $x$ that occurs in $T_{i}$, $deg(x)$ is the sum of the number of co-occurrences of $x$ with all the $ContW (T_i)$. For example, let's say we have the following  sentence  ``Feature extraction is not that complex. There are many algorithms available that can help you with feature extraction. Rapid Automatic Key Word Extraction is one of those". Unique content words of the given sentence are: `feature, extraction, complex, algorithms, available, help, rapid, automatic, keyword'. Then word degree of `feature' is (self co-occurrence $2$ times)+(co-occurrence with extraction $2$ times)=$4$. Freq(feature) is $2$ and hence $S_{wd}$(feature)=$4/2$ =$2$.

\begin{align}
   S_{wd}(x) = \frac{deg(x)}{freq(x)} 
    \label{eq:wordscor}
\end{align}\

As per Equation~\ref{eq:wordscor}, all content words are treated equally while computing their word-degree score. However, a content word related to disaster should get a higher score compared to others~\cite{garg2023ontodsumm}. To incorporate this, we modify~\footnote{We decide the modification factor on the basis of the experimental evaluation.} Equation~\ref{eq:wordscor} as follows.

\begin{equation}
S_{wd}(x) = \begin{cases}
             \frac{deg(x)}{freq(x)}, if x \notin ontology \\
             \frac{2.deg(x)}{freq(x)},otherwise.
       \end{cases} \quad
       \label{eq:modwordscor}
\end{equation}

To identify the disaster-specific words in ${Kp}{_j}^{i}$, we utilize the disaster domain knowledge from ontology~\footnote{There are several disaster-specific ontologies are available~\cite{Moi2016ontology, sermet2019towards, yahya2020ontology}, we choose Empathi~\cite{gaur2019empathi} as it provides the maximum number of disaster-specific keywords.}~\cite{gaur2019empathi}. Further, we calculate $Sc(T_{i}, {Kp}{_j}^{i})$ with this modification. Finally, we refer to that key-phrase as the final key-phrase out of all potential key-phrases, which has maximum $Sc(T_{i}, {Kp}{_j}^{i})$ score.  
The score of the final key-phrase is computed by $MaxSc({Kp}(T_i))$ as shown in Equation~\ref{eq:Maxscor}. We repeat this for all $T_{i} \in T$ to extract the key-phrases of $T$. 

\begin{align}
   MaxSc({Kp}(T_i)) = \underset{\ j \in [1, o_i]}{\operatorname{\argmax}} (Sc(T_{i}, {Kp}{_j}^{i}))
    \label{eq:Maxscor} 
\end{align}\

\par In order to understand the performance of DRAKE, we validate it by an annotator who manually identifies the key-phrases of tweets on the basis of his/her understanding of the tweet text. We consider annotated key-phrases as ground truth key-phrases of the tweets. We, then, compare extracted key-phrase by DRAKE with the ground truth key-phrase for each tweet using two metrics discussed by DeYoung et al.~\cite{deyoung2020eraser}, such as IOU (F1-score) and Jaccard Similarity score for $D_1-D_{12}$ datasets. Our results in Table~\ref{table:result_drake} indicate that the performance of DRAKE ranges between $0.25-0.38$ for IOU (F1-score) and $0.39-0.63$ for Jaccard Similarity across the datasets. Our results show that DRAKE outperforms existing supervised and unsupervised approaches (which discuss in detail in Section~\ref{s:expident}).

\begin{table}[ht]
    \centering\caption{Table shows the IOU (F1-score) and Jaccard Similarity scores for the identified key-phrase using DRAKE for $12$ disaster datasets.}
    \label{table:result_drake}
    \resizebox{\linewidth}{!}{\begin{tabular}{ccc|ccc}
        \hline
        {\bf Dataset} & \textbf{IOU (F1-score)} & \textbf{Jaccard Similarity} & {\bf Dataset}  & \textbf{IOU (F1-score)} & \textbf{Jaccard Similarity}\\ 
                      & \textbf{Score} & \textbf{Score} & & \textbf{Score} & \textbf{Score}\\ \hline

            $D_1$ & 0.3160 & 0.4423 & $D_7$ & 0.3029 & 0.4333 \\
            $D_2$ & 0.3174 & 0.4495 & $D_8$ & 0.2509 & 0.3850 \\
            $D_3$ & 0.3056 & 0.4272 & $D_9$ & 0.2809 & 0.4092 \\
            $D_4$ & 0.3206 & 0.4672 & $D_{10}$ & 0.2627 & 0.3856 \\
            $D_5$ & 0.3336 & 0.4086 & $D_{11}$ & 0.3835 & 0.6320 \\
            $D_6$ & 0.2919 & 0.4234 & $D_{12}$ & 0.2978 & 0.4629 \\\hline
  \end{tabular}}
\end{table}

\subsection{Key-phrase Encoder and Integrator} \label{s:w2v}
\par We generate the embedding of the tweet key-phrase, $KE_{T_i}$, as the average of each word embedding of the key-phrase of $T_i$. To generate the word embedding, we use a pre-trained Word2Vec model, which was trained on  $52$ million disaster-related tweets~\cite{imran2016twitter}. We, then, concatenate $KE_{T_i}$ and $TE_{T_i}$ as the final output and pass the concatenated embedding through a dense layer to calculate the salience score of $T_i$. To update the weights of the neural network, we use \textit{Binary Cross entropy}  as the loss function as shown in Equation \ref{eq:crs}.

\begin{equation}
\mathrm{Loss} = - \frac{1}{\mathrm{N}} \sum_{i=1}^{\mathrm{N}} y_i \cdot \mathrm{log}\; {f(TE_{T_i}\parallel KE_{T_i})} + (1-y_i) \cdot \mathrm{log}\; (1-{f(TE_{T_i}\parallel KE_{T_i})}) 
\label{eq:crs} 
\end{equation}

where, $f$ is the Sigmoid activation function and $y_i$ is the target value.

\subsection{Summary Tweet Selector} \label{s:sts} 
\par We aim to create a summary by selecting the $L$ tweets from $T$ such that maximizes the information coverage of $T$ and diversity in the final summary. While selecting the tweet with the maximum salience score ensures information coverage with respect to the disaster, selection of the tweet with minimum similarity with the existing summary tweets ensures maximum diversity in summary~\cite{chakraborty2019tweet,chakraborty2017network}. Therefore, we iteratively select tweets in the descending order of their salience score such that each tweet has at least a threshold $\lambda_{salience}$ score. The pseudo-code of the tweet selection process is shown in Algorithm~\ref{a:alg_1}.\\

\begin{walgorithm}
    \setstretch{1.00}
    \textbf{Input:} A set of tweets, $T$, with salience score greater than $\lambda_{salience}$ and ordered by non-increasing score.\\
    \textbf{Output:} Summary $S$ consist of a subset of tweets.  \\
    \Begin
    {%
        {$S^{'} \gets \{\}$\\ $i \gets 1$}\\
        \While{$S^{'}.length < L$ and $t \gets T[i]$}
        {
            \uIf{$S^{'}.isEmpty()$}
            {
                $S^{'}.append(t)$
            }
            \Else
            {
             \If{$Sim(t, S^{'}) < \lambda_{similarity}$}
                {
                    $S^{'}.append(t)$
                }   
            }
            $i \gets i +1$\\
        }
    }
    Return $S \gets S^{'}$.
    
    \caption{Tweet selection to obtain summary, $S$, by incrementally updating the preceding summary, $S^{'}$.}\label{a:alg_1}
\end{walgorithm}

\section{Experiments}\label{s:exp}
\par In this Section, we initially discuss the existing state-of-the-art disaster summarization approaches, which we use as baselines, followed by our experiments, results, and discussions.

\subsection{Baselines}\label{s:baseline}
\par We compare \textit{IKDSumm} with the following state-of-the-art summarization approaches:  

\begin{enumerate}
    \item \textit{$B_1$}: Rudra et al.~\cite{rudra2019summarizing} propose a graph-based summarization framework where they select tweets on the basis of the information covered by the constituent words of the tweet.

    
    \item \textit{$B_2$}: Dutta et al.~\cite{dutta2018ensemble} propose an ensemble graph-based summarization approach where they integrate the summary of $9$ existing summarization algorithms in a graph to select a tweet on the basis of different attributes, such as length, informativeness, and centrality scores, into the summary.
    
    
    \item \textit{$B_3$}: Rudra et al.~\cite{rudra2018identifying} propose a sub-event-based summarization approach where they select those tweets into the summary that provides maximum information coverage of each sub-event.
    
    
    \item \textit{$B_4$}: Nguyen et al.~\cite{nguyen2022towards} select those tweets into a summary, which has the key-phrases with maximum information.
    
    \item \textit{$B_5$}: Garg et al.~\cite{garg2023ontodsumm} propose an ontology-based pseudo-relevance feedback approach to identify categories of a disaster event and thereby, select the most representative tweets from each category into the summary.
    
    
    \item \textit{$B_6$}: Garg et al.~\cite{Garg2022Entropy} selects tweets into a summary, which provides maximum entropy and diversity in summary.

    \item \textit{$B_7$}: Dusart et al.~\cite{dusart2023tssubert} predict tweet salience score by integrating BERT model~\cite{liu2019text} and vocabulary frequency of an event to generate the summary. 
    

    \item \textit{$B_8$}: Li et al.~\cite{li2021twitter} propose a GCN-based summarization framework to identify the most important tweets of a disaster event to create a summary.
    
\end{enumerate}

\begin{table*}[htbp]
    \centering 
    \caption{Table shows F1-score of ROUGE-$1$, $2$, and $L$ score of \textit{IKDSumm} and baselines on $D_1-D_{12}$ datasets.}
    \label{table:Result1}
    \resizebox{\textwidth}{!}{\begin{tabular}{cccccccccc} \hline
    
        \textbf{Dataset} & \textbf{Approach} & \textbf{ROUGE-1} & \textbf{ROUGE-2} & \textbf{ROUGE-L} & \textbf{Dataset} & \textbf{Approach} & \textbf{ROUGE-1} & \textbf{ROUGE-2} & \textbf{ROUGE-L} \\ \cline{3-5} \cline{8-10}
        
        &  & \textbf{F1-score} & \textbf{F1-score} & \textbf{F1-score} &  &  & \textbf{F1-score} & \textbf{F1-score} & \textbf{F1-score} \\ \hline

                 & $IKDSumm $  & \bf0.57 & \bf0.29 & 0.42 &             & $IKDSumm $      & \bf0.55 & \bf0.20 & \bf0.27 \\
                 & $B_1$       & 0.53 & 0.26 & 0.33 &                   & $B_1$           & 0.48 & 0.13 & 0.22 \\
                 & $B_2$       & 0.52 & 0.22 & 0.29 &                   & $B_2$           & 0.47 & 0.14 & 0.22 \\ 
                 & $B_3$       & 0.48 & 0.20 & 0.27 &                   & $B_3$           & 0.44 & 0.12 & 0.22 \\ 
        ${D_1}$  & $B_4$       & 0.49 & 0.20 & 0.29 &          ${D_7}$  & $B_4$           & 0.50 & 0.16 & 0.25 \\ 
                 & $B_5$       & 0.54 & 0.23 & 0.29 &                   & $B_5$           & 0.51 & 0.17 & 0.26 \\
                 & $B_6$       & 0.55 & 0.27 & \bf0.44 &                & $B_6$           & 0.52 & 0.17 & 0.24 \\ 
                 & $B_7$       & 0.53 & 0.21 & 0.36 &                   & $B_7$           & 0.50 & 0.15 & 0.24 \\ 
                 & $B_8$       & 0.52 & 0.27 & 0.32 &                   & $B_8$           & 0.50 & 0.16 & 0.25 \\ \hline

                 & $IKDSumm $  & \bf0.48 & \bf0.19 & \bf0.29 &          & $IKDSumm $           & \bf0.54 & \bf0.20 & \bf0.29 \\
                 & $B_1$       & 0.33 & 0.14 & 0.24 &                   & $B_1$                & 0.45 & 0.13 & 0.23 \\
                 & $B_2$       & 0.37 & 0.16 & 0.25 &                   & $B_2$                & 0.46 & 0.14 & 0.24 \\
                 & $B_3$       & 0.41 & 0.12 & 0.24 &                   & $B_3$                & 0.44 & 0.14 & 0.23 \\
        ${D_2}$  & $B_4$       & 0.38 & 0.13 & 0.25 &          ${D_8}$  & $B_4$                & 0.50 & 0.17 & 0.25 \\
                 & $B_5$       & 0.46 & 0.18 & 0.27 &                   & $B_5$                & 0.52 & 0.19 & 0.27 \\
                 & $B_6$       & 0.44 & 0.17 & 0.25 &                   & $B_6$                & 0.52 & 0.14 & 0.26 \\ 
                 & $B_7$       & 0.40 & 0.14 & 0.26 &                   & $B_7$                & 0.51 & 0.16 & 0.26 \\
                 & $B_8$       & 0.42 & 0.16 & 0.24 &                   & $B_8$                & 0.50 & 0.17 & 0.25 \\ \hline

                 & $IKDSumm $  & \bf0.48 & \bf0.22 & \bf0.32 &          & $IKDSumm $           & \bf0.57 & \bf0.18 & \bf0.25 \\
                 & $B_1$       & 0.36 & 0.14 & 0.29 &                   & $B_1$                & 0.20 & 0.04 & 0.20  \\ 
                 & $B_2$       & 0.40 & 0.18 & 0.27 &                   & $B_2$                & 0.47 & 0.14 & 0.21 \\
                 & $B_3$       & 0.44 & 0.17 & 0.24 &                   & $B_3$                & 0.45 & 0.11 & 0.21 \\
        ${D_3}$  & $B_4$       & 0.41 & 0.15 & 0.29 &          ${D_9}$  & $B_4$                & 0.44 & 0.10 & 0.20 \\
                 & $B_5$       & 0.47 & 0.20 & 0.30 &                   & $B_5$                & 0.54 & 0.17 & 0.24 \\ 
                 & $B_6$       & 0.43 & 0.16 & 0.29 &                   & $B_6$                & 0.51 & 0.16 & 0.24 \\
                 & $B_7$       & 0.44 & 0.17 & 0.30 &                   & $B_7$                & 0.48 & 0.12 & 0.21 \\ 
                 & $B_8$       & 0.25 & 0.06 & 0.17 &                   & $B_8$                & 0.51 & 0.16 & 0.22 \\ \hline

                 & $IKDSumm $  & \bf0.47 & \bf0.20 & \bf0.29 &          & $IKDSumm $           & \bf0.56 & \bf0.19 & \bf0.25 \\
                 & $B_1$       & 0.32 & 0.13 & 0.24 &                   & $B_1$                & 0.19 & 0.04 & 0.18 \\
                 & $B_2$       & 0.41 & 0.16 & 0.26 &                   & $B_2$                & 0.48 & 0.10 & 0.20 \\
                 & $B_3$       & 0.45 & 0.17 & 0.25 &                   & $B_3$                & 0.50 & 0.12 & 0.22 \\
        ${D_4}$  & $B_4$       & 0.36 & 0.13 & 0.27 &       ${D_{10}}$  & $B_4$                & 0.52 & 0.12 & 0.21 \\
                 & $B_5$       & 0.47 & 0.19 & 0.28 &                   & $B_5$                & 0.55 & 0.17 & 0.24 \\
                 & $B_6$       & 0.43 & 0.16 & 0.25 &                   & $B_6$                & 0.52 & 0.13 & 0.24 \\
                 & $B_7$       & 0.41 & 0.14 & 0.27 &                   & $B_7$                & 0.50 & 0.12 & 0.21 \\
                 & $B_8$       & 0.44 & 0.17 & 0.27 &                   & $B_8$                & 0.46 & 0.10 & 0.20 \\ \hline

                 & $IKDSumm $  & \bf0.61 & \bf0.36 & \bf0.37 &          & $IKDSumm $           & \bf0.55 & \bf0.18 & \bf0.25 \\
                 & $B_1$       & 0.44 & 0.23 & 0.29 &                   & $B_1$                & 0.47 & 0.15 & 0.22 \\
                 & $B_2$       & 0.51 & 0.23 & 0.28 &                   & $B_2$                & 0.42 & 0.09 & 0.20 \\
                 & $B_3$       & 0.53 & 0.25 & 0.27 &                   & $B_3$                & 0.43 & 0.11 & 0.19 \\
        ${D_5}$  & $B_4$       & 0.38 & 0.16 & 0.32 &       ${D_{11}}$  & $B_4$                & 0.45 & 0.14 & 0.22 \\
                 & $B_5$       & 0.59 & 0.35 & 0.35 &                   & $B_5$                & 0.53 & 0.17 & 0.25 \\
                 & $B_6$       & 0.54 & 0.25 & 0.33 &                   & $B_6$                & 0.47 & 0.13 & 0.21 \\
                 & $B_7$       & 0.49 & 0.17 & 0.30 &                   & $B_7$                & 0.47 & 0.15 & 0.23 \\ 
                 & $B_8$       & 0.42 & 0.20 & 0.31 &                   & $B_8$                & 0.46 & 0.14 & 0.19 \\ \hline

                 & $IKDSumm $  & \bf0.59 & \bf0.25 & \bf0.32 &          & $IKDSumm $           & \bf0.57 & \bf0.25 & \bf0.34 \\
                 & $B_1$       & 0.49 & 0.22 & 0.29 &                   & $B_1$                & 0.49 & 0.18 & 0.30 \\
                 & $B_2$       & 0.48 & 0.18 & 0.25 &                   & $B_2$                & 0.49 & 0.17 & 0.29 \\
                 & $B_3$       & 0.52 & 0.21 & 0.23 &                   & $B_3$                & 0.49 & 0.17 & 0.28 \\
        ${D_6}$  & $B_4$       & 0.53 & 0.21 & 0.29 &      ${D_{12}}$   & $B_4$                & 0.48 & 0.18 & 0.29 \\
                 & $B_5$       & 0.57 & 0.24 & 0.30 &                   & $B_5$                & 0.56 & 0.25 & 0.33 \\
                 & $B_6$       & 0.55 & 0.21 & 0.27 &                   & $B_6$                & 0.49 & 0.21 & 0.29 \\ 
                 & $B_7$       & 0.55 & 0.23 & 0.28 &                   & $B_7$                & 0.50 & 0.18 & 0.29 \\
                 & $B_8$       & 0.52 & 0.20 & 0.27 &                   & $B_8$                & 0.51 & 0.21 & 0.31 \\ \hline 
    \end{tabular}}
\end{table*}

\subsection{Experimental Setup}\label{s:expsetup}
\par We use the following hyper-parameters in the \textit{IKDSumm} experiments. For each dense layer, we use a ReLu activation function. In addition, we use a Dropout layer as proposed by~\cite{srivastava2014dropout} after each fully connected layer with a probability of $0.5$. We train for $3$ epochs with Adam optimizer~\cite{kingma2014adam} with parameters $\beta_1$ = $0.9$ and $\beta_2$ = $0.999$, batch size of $128$ with a Mean Squared Error objective function. We use $\lambda_{salience}=0.2$ and $\lambda_{similarity}=0.3$, which was set experimentally by Dusart et al.~\cite{dusart2023tssubert}. We consider cosine Similarity~\cite{lahitani2016cosine} to calculate the similarity, and we follow Dusart et al.~\cite{dusart2023tssubert}, to decide our training datasets, i.e., which trains the model on all the datasets except the test dataset.

\subsection{Comparison with Existing Research Works}\label{s:result}
\par We evaluate the performance of the generated summaries of \textit{IKDSumm} and baselines with the ground truth summary using ROUGE-N~\cite{lin2004rouge} score. ROUGE-N computes the score on the basis of overlapping words between the system-generated summary with the ground truth summary, and a higher value of the ROUGE score indicates a good quality summary. We use F1-score for $3$ different variants of the ROUGE-N score, i.e., N=$1$, $2$, and $L$, respectively. 

\par Our observations as shown in Table~\ref{table:Result1} indicate that \textit{IKDSumm} ensures better ROUGE-N F1-score in comparison with baselines except for $D_1$ ROUGE-L F1-score by $0.02$ where $B_6$ performers better than \textit{IKDSumm}. The improvement in summary scores of ROUGE-1 F1-score ranges from $1.78\%$ to $66.07\%$, ROUGE-2 F1-score ranges from $5.00\%$ to $78.95\%$ and ROUGE-L F1-score ranges from $3.44\%$ to $46.87\%$, respectively. The improvement is highest over the $B_1$ baseline and lowest with the $B_5$ baseline. \textit{IKDSumm} gains highest improvement for $D_5$ dataset with $0.61-0.36$ and the gain is minimum for $D_4$ dataset with $0.47-0.20$ in terms of ROUGE-1, ROUGE-2, and ROUGE-L F1-score. 


\subsection{Ablation Experiments} \label{s:ablexp}
\par To understand the effectiveness of each component of \textit{IKDSumm}, we compare and validate the performance of \textit{IKDSumm} with its different variants, which are as follows:

\begin{table}[ht]
    \centering 
    \caption{Table shows F1-score of ROUGE-$1$, $2$, and $L$ for $5$ different variants of our proposed \textit{IKDSumm}, i.e., \textit{IKDSumm-A}, \textit{IKDSumm-B}, \textit{IKDSumm-C}, \textit{IKDSumm-D}, and \textit{IKDSumm-E} with \textit{IKDSumm} on $D_1-D_{12}$ disasters.}
    \label{table:cat_train}
    \resizebox{\textwidth}{!}{\begin{tabular}{clcccclccc} \hline
    
        \textbf{Dataset} & \textbf{Approach} & \textbf{ROUGE-1} & \textbf{ROUGE-2} & \textbf{ROUGE-L}  & \textbf{Dataset} & \textbf{Approach} & \textbf{ROUGE-1} & \textbf{ROUGE-2} & \textbf{ROUGE-L}  \\ \cline{3-5} \cline{08-10}
        &  & \textbf{F1-score} & \textbf{F1-score} & \textbf{F1-score} & &  & \textbf{F1-score} & \textbf{F1-score} & \textbf{F1-score} \\ \hline

                 & $IKDSumm$       & \bf0.57 & \bf0.29 & \bf0.42 & & $IKDSumm$   & \bf0.55 & \bf0.20 & \bf0.27 \\ 
                 & $IKDSumm-A$     & 0.49 & 0.19 & 0.40 &          & $IKDSumm-A$ & 0.45 & 0.11 & 0.22 \\ 
        ${D_1}$  & $IKDSumm-B$     & 0.53 & 0.21 & 0.36 & ${D_7}$  & $IKDSumm-B$ & 0.50 & 0.15 & 0.24 \\ 
                 & $IKDSumm-C$     & 0.51 & 0.19 & 0.36 &          & $IKDSumm-C$ & 0.44 & 0.12 & 0.23 \\ 
                 & $IKDSumm-D$     & 0.54 & 0.23 & 0.38 &          & $IKDSumm-D$ & 0.53 & 0.20 & 0.26 \\ 
                 & $IKDSumm-E$     & 0.54 & 0.23 & 0.39 &          & $IKDSumm-E$ & 0.54 & 0.20 & 0.27 \\ \hline

                 & $IKDSumm$       & \bf0.48 & \bf0.19 & \bf0.29 & & $IKDSumm$   & \bf0.54 & \bf0.20 & \bf0.29 \\ 
                 & $IKDSumm-A$     & 0.37 & 0.16 & 0.22 &          & $IKDSumm-A$ & 0.47 & 0.13 & 0.24 \\ 
        ${D_2}$  & $IKDSumm-B$     & 0.40 & 0.14 & 0.26 & ${D_8}$  & $IKDSumm-B$ & 0.51 & 0.16 & 0.26 \\ 
                 & $IKDSumm-C$     & 0.39 & 0.12 & 0.22 &          & $IKDSumm-C$ & 0.50 & 0.15 & 0.24 \\ 
                 & $IKDSumm-D$     & 0.42 & 0.15 & 0.24 &          & $IKDSumm-D$ & 0.52 & 0.19 & 0.28 \\ 
                 & $IKDSumm-E$     & 0.42 & 0.16 & 0.25 &          & $IKDSumm-E$ & 0.53 & 0.19 & 0.28 \\ \hline

                 & $IKDSumm$       & \bf0.48 & \bf0.22 & \bf0.32 & & $IKDSumm$   & \bf0.57 & \bf0.18 & \bf0.25 \\ 
                 & $IKDSumm-A$     & 0.41 & 0.15 & 0.25 &          & $IKDSumm-A$ & 0.47 & 0.10 & 0.20 \\ 
        ${D_3}$  & $IKDSumm-B$     & 0.44 & 0.17 & 0.30 & ${D_9}$  & $IKDSumm-B$ & 0.48 & 0.12 & 0.21 \\ 
                 & $IKDSumm-C$     & 0.42 & 0.14 & 0.26 &          & $IKDSumm-C$ & 0.47 & 0.12 & 0.20 \\ 
                 & $IKDSumm-D$     & 0.46 & 0.20 & 0.30 &          & $IKDSumm-D$ & 0.54 & 0.18 & 0.24 \\ 
                 & $IKDSumm-E$     & 0.47 & 0.21 & 0.30 &          & $IKDSumm-E$ & 0.56 & 0.18 & 0.25 \\ \hline

                 & $IKDSumm$       & \bf0.47 & \bf0.20 & \bf0.29 & & $IKDSumm$   & \bf0.56 & \bf0.19 & \bf0.25 \\ 
                 & $IKDSumm-A$     & 0.38 & 0.11 & 0.25 &          & $IKDSumm-A$ & 0.47 & 0.11 & 0.18 \\ 
        ${D_4}$  & $IKDSumm-B$     & 0.41 & 0.14 & 0.27 &${D_{10}}$& $IKDSumm-B$ & 0.50 & 0.12 & 0.21 \\  
                 & $IKDSumm-C$     & 0.40 & 0.12 & 0.26 &          & $IKDSumm-C$ & 0.47 & 0.12 & 0.20 \\ 
                 & $IKDSumm-D$     & 0.45 & 0.19 & 0.28 &          & $IKDSumm-D$ & 0.55 & 0.17 & 0.24 \\ 
                 & $IKDSumm-E$     & 0.46 & 0.19 & 0.28 &          & $IKDSumm-E$ & 0.55 & 0.18 & 0.25 \\ \hline

                 & $IKDSumm$       & \bf0.61 & \bf0.36 & \bf0.37 & & $IKDSumm$   & \bf0.55 & \bf0.18 & \bf0.25 \\ 
                 & $IKDSumm-A$     & 0.46 & 0.16 & 0.27 &          & $IKDSumm-A$ & 0.48 & 0.13 & 0.20 \\ 
        ${D_5}$  & $IKDSumm-B$     & 0.49 & 0.17 & 0.30 &${D_{11}}$& $IKDSumm-B$ & 0.47 & 0.15 & 0.23 \\  
                 & $IKDSumm-C$     & 0.48 & 0.17 & 0.27 &          & $IKDSumm-C$ & 0.46 & 0.14 & 0.23 \\ 
                 & $IKDSumm-D$     & 0.57 & 0.33 & 0.34 &          & $IKDSumm-D$ & 0.52 & 0.16 & 0.23 \\ 
                 & $IKDSumm-E$     & 0.58 & 0.34 & 0.35 &          & $IKDSumm-E$ & 0.52 & 0.17 & 0.25 \\ \hline
                 
                 & $IKDSumm$       & \bf0.59 & \bf0.25 & \bf0.32 & & $IKDSumm$   & \bf0.57 & \bf0.25 & \bf0.34 \\ 
                 & $IKDSumm-A$     & 0.53 & 0.21 & 0.25 &          & $IKDSumm-A$ & 0.47 & 0.14 & 0.24 \\ 
        ${D_6}$  & $IKDSumm-B$     & 0.55 & 0.23 & 0.28 &${D_{12}}$& $IKDSumm-B$ & 0.50 & 0.18 & 0.29 \\ 
                 & $IKDSumm-C$     & 0.51 & 0.17 & 0.25 &          & $IKDSumm-C$ & 0.46 & 0.15 & 0.24 \\ 
                 & $IKDSumm-D$     & 0.57 & 0.23 & 0.29 &          & $IKDSumm-D$ & 0.55 & 0.23 & 0.33 \\ 
                 & $IKDSumm-E$     & 0.57 & 0.24 & 0.30 &          & $IKDSumm-E$ & 0.55 & 0.23 & 0.33 \\ \hline
    \end{tabular} }
\end{table}

\begin{enumerate}
    \item \textbf{IKDSumm-A}, we employ only the DistilBERT model to predict tweet salience score and then select the highest salience score tweets into the summary. 
    
    \item \textbf{IKDSumm-B}, we combine tweet embedding with the tweet context, which measures using the whole vocabulary related to the disaster instead of key-phrase~\cite{dusart2023tssubert}. 
    
    \item \textbf{IKDSumm-C}, we combine tweet embedding with the importance of that tweet which we measure using TF-IDF. 
    
    \item \textbf{IKDSumm-D}, we combine tweet embedding with key-phrase embedding, where we identify the key-phrase using a supervised approach proposed by Nguyen et al.~\cite{nguyen2022towards}.

    \item \textbf{IKDSumm-E}, we combine tweet embedding with key-phrase embedding, where we identify the key-phrase using RAKE~\cite{rose2010automatic}.
\end{enumerate}

\par We show our comparison results of \textit{IKDSumm} with the variants on $D_1-D_{12}$ in Table~\ref{table:cat_train}. Our results indicate that the inclusion of DRAKE in \textit{IKDSumm} provides an increase of $1.75-24.59\%$, $4.00-55.56\%$, and $2.94-29.41\%$ in ROUGE-1, 2, and L F1-score, respectively, when compared to all the variants of \textit{IKDSumm} which indicates the requirement of the utilization of domain knowledge to determine key-phrases. Additionally, we observe that the performance gain of \textit{IKDSumm} is the highest  over \textit{IKDSumm-A}, which only uses the DistilBERT model without combining the tweet's key-phrases ( $10.17-24.59\%$, $15.79-55.56\%$, and $8.33-29.41\%$ in ROUGE-1, 2, and L F1-score, respectively), which indicates the need to include key-phrases in disaster tweet summarization. Additionally, we observe that \textit{IKDSumm} performs better than \textit{IKDSumm-D} as $1.75-12.50\%$, $4.00-20.69\%$, and $2.94-17.24\%$ in ROUGE-1, 2, and L F1-score, respectively, although \textit{IKDSumm-D} uses a supervised approach for key-phrase identification. 


\subsection{Extraction of Key-phrase of a Tweet} \label{s:expident}
\par In this Subsection, we evaluate the effectiveness of the proposed unsupervised key-phrase identification approach, i.e., DRAKE, with existing supervised and unsupervised approaches. We select RAKE~\cite{rose2010automatic} as an unsupervised approach, and Nguyen et al.~\cite{nguyen2022towards} as a supervised approach on the basis of its performance recency. While Nguyen et al.~\cite{nguyen2022towards} propose a BERT-based multi-task learning approach to extract key-phrases from tweets, RAKE relies on word co-occurrences and frequency to identify key-phrases without any training. For our experiment, we consider all the tweets for $D_1-D_{12}$ disaster events. For our ground truth, we ask $3$ annotators to manually annotate a given phrase as key-phrase or not, and we consider a phrase as key-phrase if a majority of the annotators agree on that. As shown in Table~\ref{table:result_compare}, DRAKE performs better than RAKE, and Nguyen et al.~\cite{nguyen2022towards} by $5.10-25.78\%$ and $4.02-27.83\%$ in IOU (F1-score) and $7.34-26.53\%$ and $2.33-20.27\%$, respectively, in Jaccard Similarity score. Therefore, as DRAKE utilizes the domain knowledge of disasters, it performs better than both  RAKE, and Nguyen et al.~\cite{nguyen2022towards}. Additionally, DRAKE does not require human intervention.


\begin{table}[ht]
    \centering\caption{Table shows the IOU (F1-score) and Jaccard Similarity score for the identified key-phrases using Nguyen et al.~\cite{nguyen2022towards}, RAKE, and DRAKE for $12$ disaster datasets.}
    \label{table:result_compare}
    \resizebox{\linewidth}{!}{\begin{tabular}{cP{0.2\linewidth}ccP{0.2\linewidth}cc}
        \hline
        {\bf Dataset} & \multicolumn{3}{c}{\textbf{IOU (F1-score) Score}} & \multicolumn{3}{c}{\textbf{Jaccard Similarity Score}} \\ \cline{2-7}
                      & {\bf Nguyen et al.~\cite{nguyen2022towards}} & {\bf RAKE} & {\bf DRAKE} & {\bf Nguyen et al.~\cite{nguyen2022towards}} & {\bf RAKE} & {\bf DRAKE} \\ \hline
            
            $D_1$ & 0.2734 & 0.2987 & \bf0.3160 & 0.4029 & 0.4058 & \bf0.4423 \\
            $D_2$ & 0.3000 & 0.2940 & \bf0.3174 & 0.4055 & 0.4037 & \bf0.4495 \\
            $D_3$ & 0.2873 & 0.2742 & \bf0.3056 & 0.3831 & 0.3682 & \bf0.4272 \\
            $D_4$ & \bf0.3206 & 0.3022 & \bf0.3206 & 0.4310 & 0.4329 & \bf0.4672 \\
            $D_5$ & 0.2591 & 0.2476 & \bf0.3336 & 0.3991 & 0.3002 & \bf0.4086 \\
            $D_6$ & 0.2794 & 0.2770 & \bf0.2919 & 0.3953 & 0.3853 & \bf0.4234 \\
            $D_7$ & 0.2907 & 0.2687 & \bf0.3029 & 0.3980 & 0.3657 & \bf0.4333 \\
            $D_8$ & 0.2048 & 0.2019 & \bf0.2509 & 0.3605 & 0.2923 & \bf0.3850 \\
            $D_9$ & 0.2312 & 0.2611 & \bf0.2809 & 0.3399 & 0.3688 & \bf0.4092 \\
         $D_{10}$ & 0.1896 & 0.2140 & \bf0.2627 & 0.3379 & 0.3032 & \bf0.3856 \\
         $D_{11}$ & 0.3258 & 0.3629 & \bf0.3835 & 0.5039 & 0.5794 & \bf0.6320 \\
         $D_{12}$ & 0.2488 & 0.2712 & \bf0.2978 & 0.3865 & 0.4105 & \bf0.4629 \\\hline
  \end{tabular}}
\end{table}

\begin{figure}
    \centering
    \begin{subfigure}[b]{0.45\linewidth}
    \includegraphics[width=\linewidth]{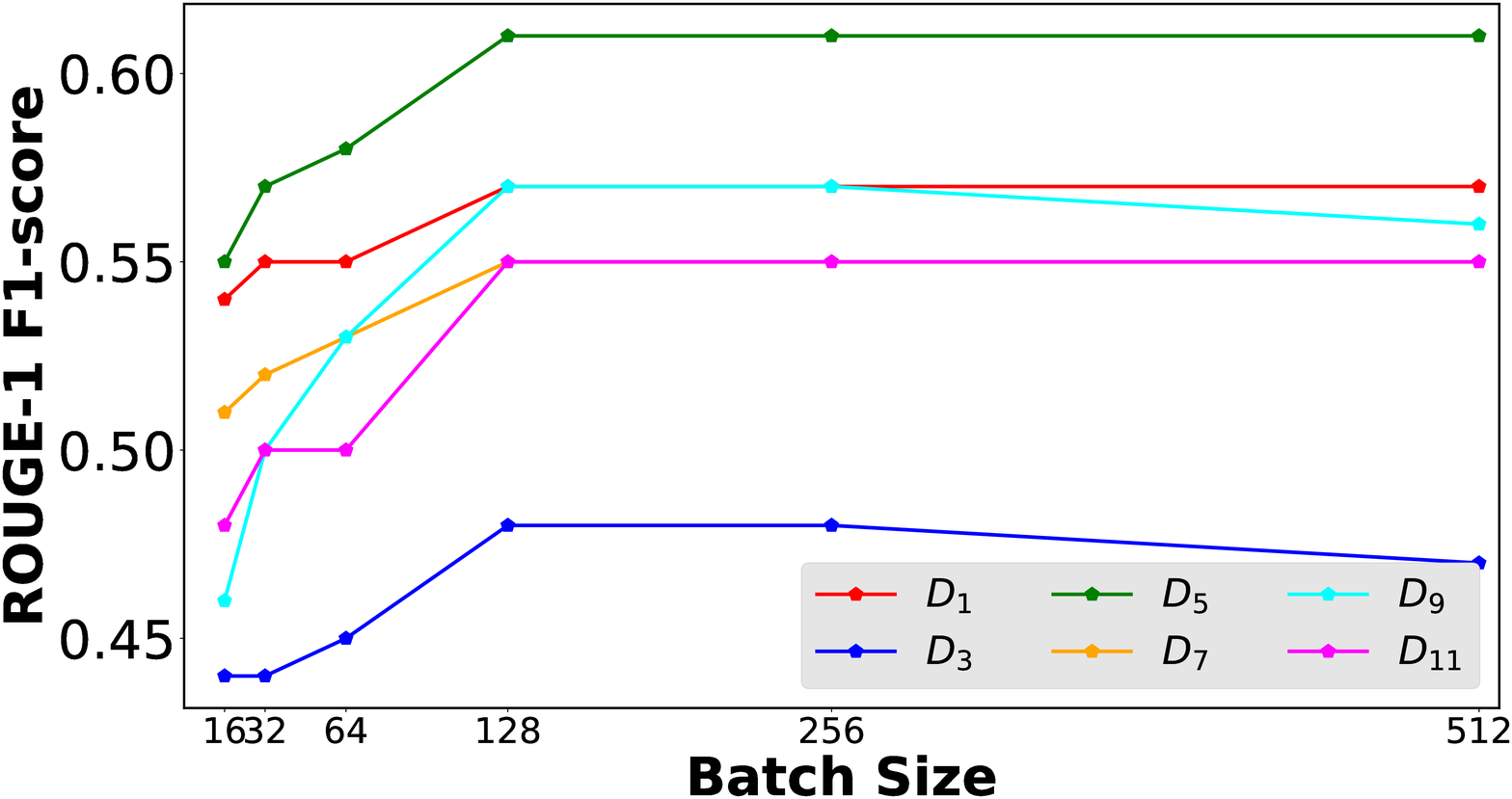}\caption{}
    \label{fig:batch}
    \end{subfigure}
    \begin{subfigure}[b]{0.45\linewidth}
    \includegraphics[width=\linewidth]{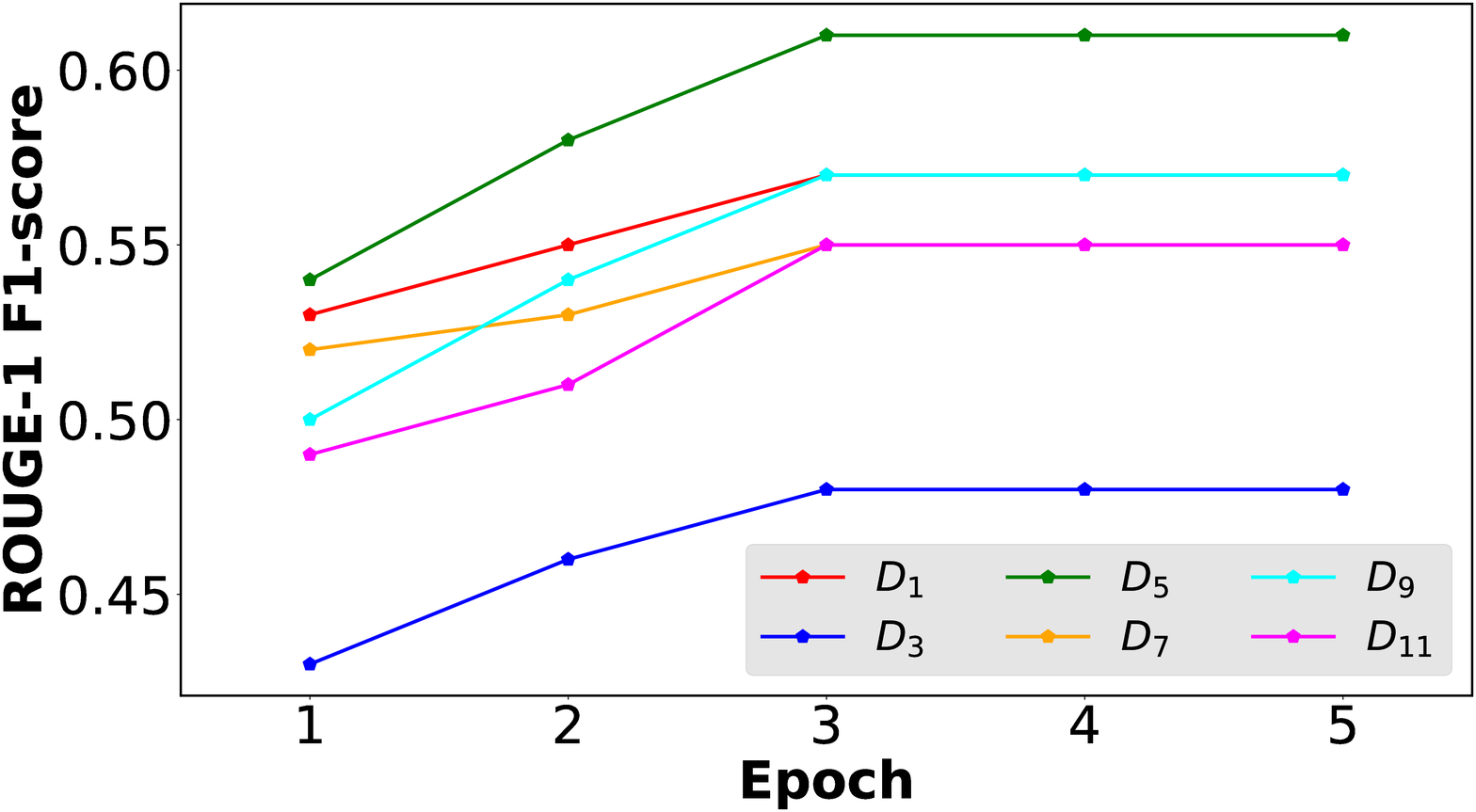}\caption{} \label{fig:epoch}
    \end{subfigure}
    \begin{subfigure}[b]{0.45\linewidth}
    \includegraphics[width=\linewidth]{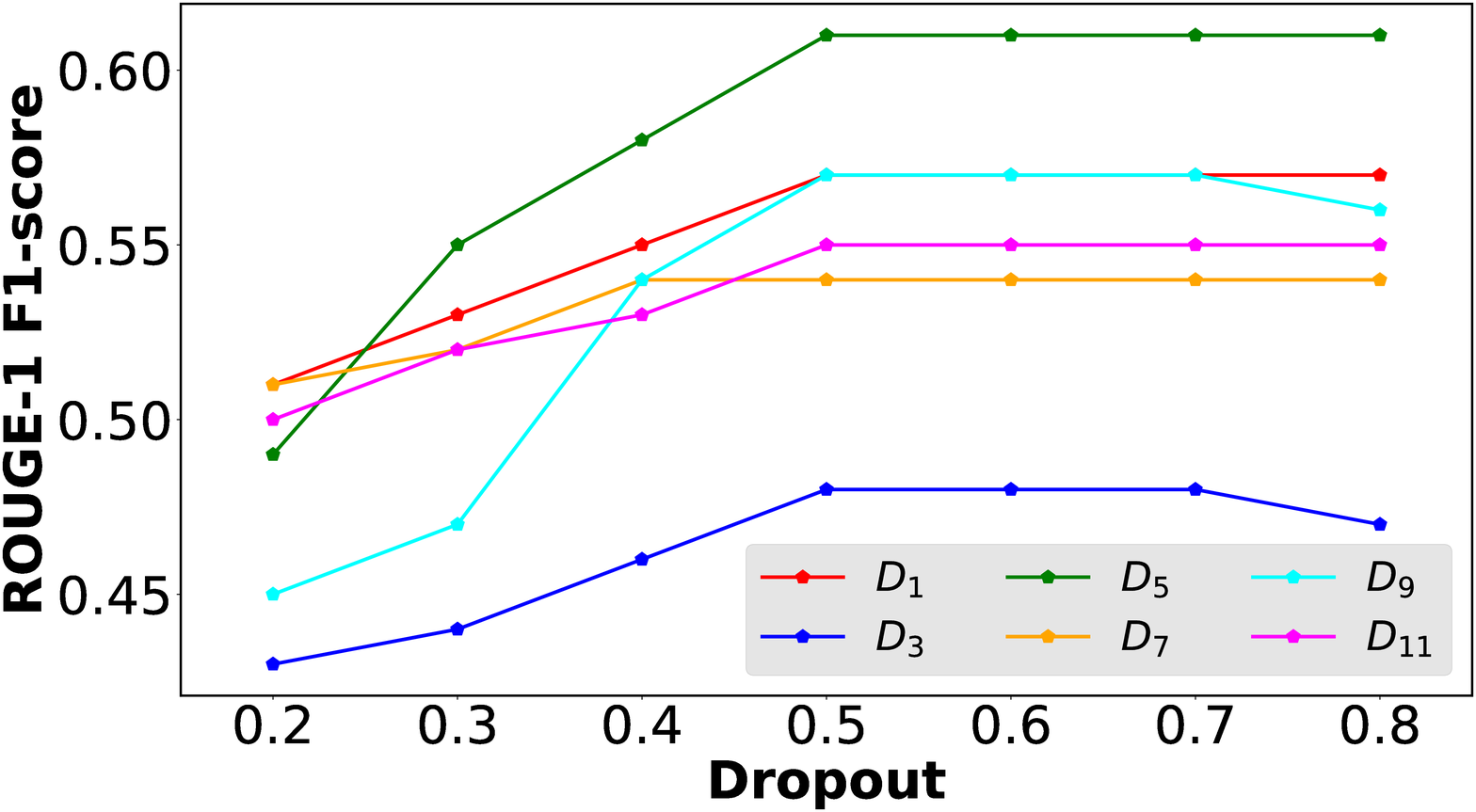} \caption{}\label{fig:dropout}
    \end{subfigure}
    \caption{Comparing the effect of change in batch size, number of epochs, and probability of Dropout layer on ROUGE-1 F1-score in Figure~\ref{fig:batch},~\ref{fig:epoch}, and \ref{fig:dropout}, respectively.} 
    \label{figure:hyperpara}
\end{figure}

\subsection{Hyperparameter Analysis} 
\par In this Subsection, we discuss how we set the values of the different hyperparameters, i.e., batch size, the number of training epochs, and Dropout layer probability for \textit{IKDSumm}. For each hyperparameter, we vary its value and calculate ROUGE-1 F1-score for each value, and then we select that value for a hyperparameter that gives the best ROUGE-1 F1-score. For the experiments, we randomly select $6$ disasters: $D_1$, $D_3$, $D_5$, $D_7$, $D_9$ and $D_{11}$. Our observations are shown in Figure~\ref{figure:hyperpara}. On the basis of our observation from Figure~\ref{fig:batch} where the batch size is changed from $16$ to $512$, we find the value of the batch size of $128$ provides the maximum ROUGE-1 F1-score for the datasets. We also observe that there is no significant increase in ROUGE-1 F1-score when batch size is increased from $128$ to $512$. Therefore, we consider $128$ as the batch size of \textit{IKDSumm}. Similarly, we observe the number of epochs as $3$ provides the maximum ROUGE-1 F1-score for \textit{IKDSumm} as shown in Figure~\ref{fig:epoch}. Additionally, to decide the probability of the Dropout layer, we vary the probability from $0.2$ to $0.8$. and calculate ROUGE-1 F1-score for \textit{IKDSumm}. Our observations, as shown in Figure~\ref{fig:dropout} indicate that increasing the Dropout layer probability from $0.5$ to $0.8$ does not improve the ROUGE-1 F1-score, and the probability of $0.2$ has the least ROUGE-1 F1-score. Therefore, we consider $0.5$ as the probability of the Dropout layer of \textit{IKDSumm} that provides maximum ROUGE-1 F1-score. 

\begin{figure}[ht]
    \centering
    \begin{subfigure}[b]{0.47\linewidth} 
    \includegraphics[width=\linewidth]{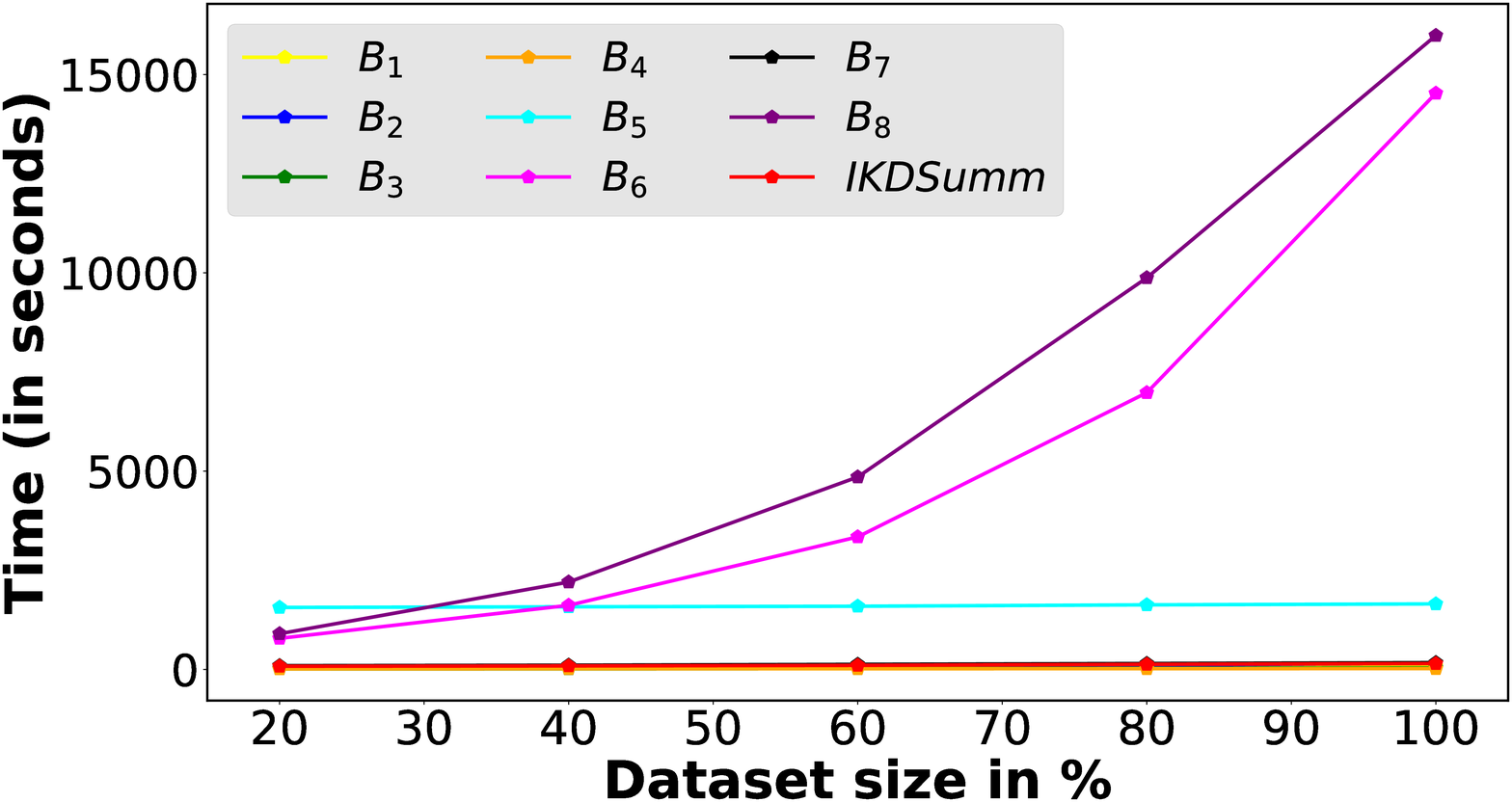}\caption{} \label{fig:testtime}
    \end{subfigure}
    \begin{subfigure}[b]{0.47\linewidth}
    \includegraphics[width=\linewidth]{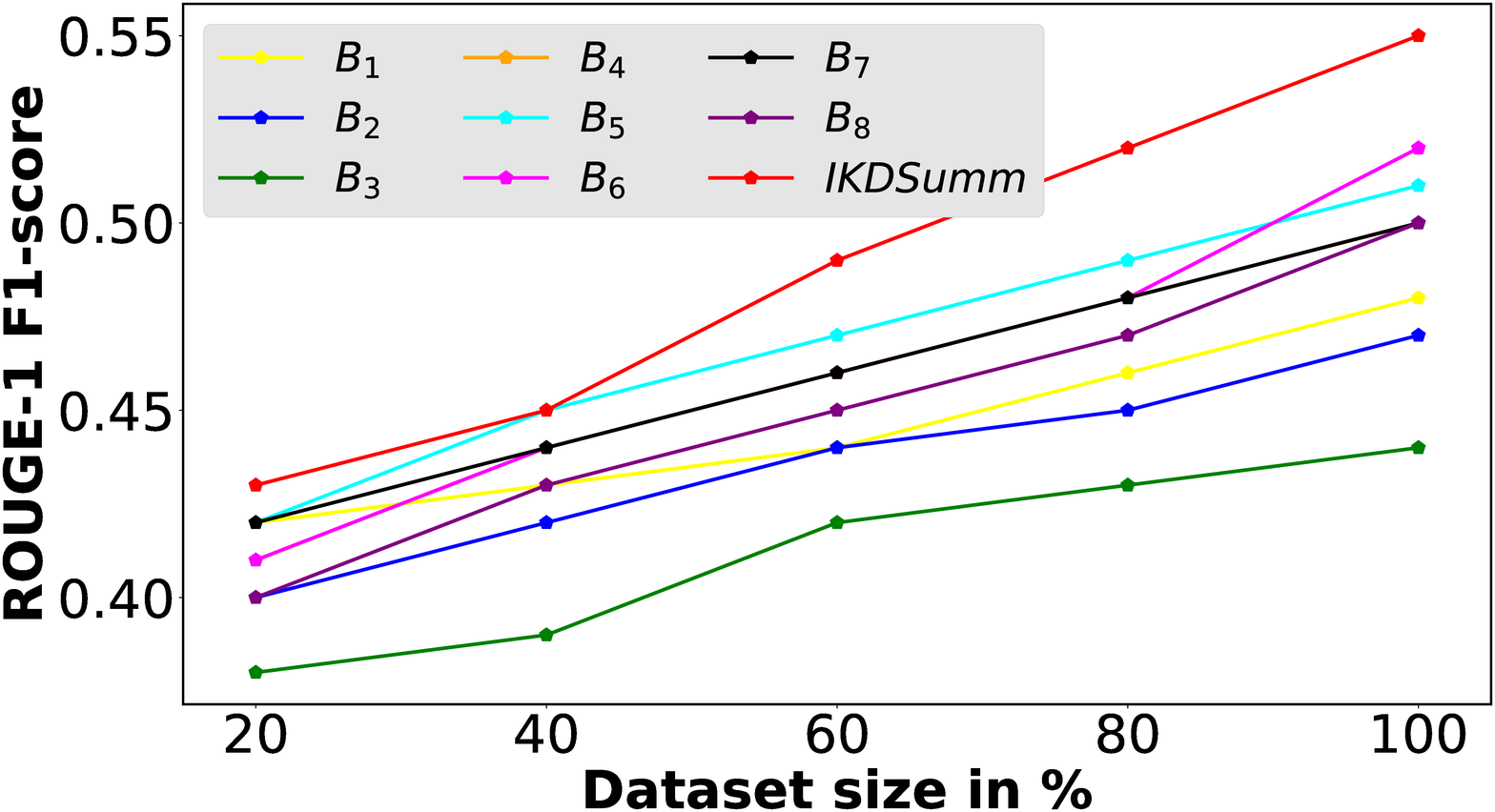}\caption{}
    \label{fig:testf1}
    \end{subfigure}
    \caption{Comparision results of execution time and ROUGE-1 F1-score of \textit{IKDSumm} with the Baselines for different dataset size in Figure~\ref{fig:testtime} and ~\ref{fig:testf1}, respectively.}
    \label{figure:performance}
\end{figure}

\subsection{Scalability Analysis} 
\par In this subsection, we investigate the scalability of \textit{IKDSumm} and the baselines by comparing their execution time and performance when the dataset size increases gradually from $20\%$ to $100\%$. For our experiment, we randomly select  the $D_7$ dataset and follow the same procedure for implementation as discussed in Section~\ref{s:expsetup}. We repeat the same experiment by varying the dataset size and then calculate ROUGE-1 F1-score and execution (time in seconds). Our observations are shown in Figure~\ref{figure:performance}. Based on our observation from Figure~\ref{fig:testtime} indicate that \textit{IKDSumm} ensures better execution time than the baselines with increases in the size of the dataset. Subsequently, we observe that execution time is highest with the $B_8$ baseline and lowest with the $B_3$ baseline. We also observe that \textit{IKDSumm} is scalable as the dataset size increases and requires around $80$ seconds for $20\%$ of the dataset and $150$ seconds for $100\%$ of the dataset. Additionally, we observe that the \textit{IKDSumm} can ensure considerably high performance in terms of ROUGE-1 F1-score than the baselines irrespective of the dataset size, as shown in Figure~\ref{fig:testf1}. Subsequently, we observe that \textit{IKDSumm} ensures the highest ROUGE-1 F1-score and lowest with the $B_3$ baseline. Therefore, we conclude that the continuous learning from the combined embedding of tweets and key-phrases ensures the high performance of \textit{IKDSumm} irrespective of the dataset size but takes more time for execution than some baselines. 


\section{Conclusions and Future Works}\label{s:con}
\par In this paper, we propose \textit{IKDSumm}, which can automatically generate a summary of the tweets related to a disaster event without any human intervention. The major novelty of \textit{IKDSumm} is the utilization of domain knowledge (ontology) in identifying key-phrases of tweets by DRAKE and integration of the key-phrases in a BERT-based approach for an effective summary generation. Therefore, \textit{IKDSumm} can fulfill the objectives of tweet summarization, such as information coverage, relevance, and diversity in summary, without any human intervention. Our experimental analysis shows that \textit{IKDSumm} can ensure $2-79\%$ higher ROUGE-N F1-score than existing state-of-the-art approaches. As a future direction, we are working on developing an automated mechanism that provides a score for every word present in the ontology on the basis of its importance to the disaster event. For example, in \textit{IKDSumm}, we currently give equal importance to every word present in the ontology. However, we intuitively believe an importance-based scoring of the ontology words would help us better rank the relevant tweets of a disaster.


\section*{CRediT authorship contribution statement}
\par Piyush Kumar Garg : Conceptualization, Methodology, Experimentation, Writing - original draft; Roshni Chakraborty : Conceptualization, Writing : review and editing, Supervision; Srishti Gupta : Conceptualization, Writing - original draft; Sourav Kumar Dandapat: Conceptualization, Writing : review and editing, Supervision.

\section*{Declaration of Competing Interest}
\par The authors have no relevant financial or non-financial interests to disclose.

\bibliographystyle{cas-model2-names}

\bibliography{reference}

\end{document}